\documentclass[runningheads]{llncs}
\usepackage{graphicx}
\usepackage{amsmath,amsfonts,amssymb} 
\usepackage{color}
\usepackage{booktabs}
\usepackage{subfig}
\usepackage[breaklinks=true,colorlinks=true,citecolor=blue]{hyperref}
\usepackage[capitalise]{cleveref}
\usepackage{tikz}
\usetikzlibrary{bayesnet,positioning,shapes.geometric}

\usepackage{amsmath,amsfonts,amssymb}
\usepackage{mathtools}
\usepackage{xparse}

\let\originalleft\left
\let\originalright\right
\renewcommand{\left}{\mathopen{}\mathclose\bgroup\originalleft}
\renewcommand{\right}{\aftergroup\egroup\originalright}

\newcommand{\cond}{\mathbin{|}}
\newcommand{\midcond}{\,\middle|\,}

\NewDocumentCommand\expec{somg}{
	\mathbb{E}%
	\IfValueT{#2}%
		{_{#2}}%
	\IfBooleanTF{#1}{
		\left[#3%
		\IfValueT{#4}%
			{\midcond #4}%
		\right]
	}{%
		[#3%
		\IfValueT{#4}%
			{\cond #4}%
		]%
	}%
}

\NewDocumentCommand\variance{somg}{
	\mathbb{V}\mathrm{ar}%
	\IfValueT{#2}%
		{_{#2}}%
	\IfBooleanTF{#1}{
		\!\left[#3%
		\IfValueT{#4}%
			{\midcond #4}%
		\right]
	}{%
		[#3%
		\IfValueT{#4}%
			{\cond #4}%
		]%
	}%
}

\def\*#1{\mathbf{#1}}

\def\bgamma{\boldsymbol{\gamma}}
\def\btheta{\boldsymbol{\theta}}

\def\bpi{\boldsymbol{\pi}}

\def\bomega{\boldsymbol{\omega}}

\NewDocumentCommand\BracketsNoStar{mg}{
	(#1%
	\IfValueT{#2}%
		{\cond #2}%
	)%
}
\NewDocumentCommand\BracketsStar{mg}{
	\left(#1%
	\IfValueT{#2}%
		{\midcond #2}%
	\right)%
}
\NewDocumentCommand\CondBrackets{smg}{
	\IfBooleanTF{#1}{%
		\BracketsStar{#2}{#3}
	}{%
		\BracketsNoStar{#2}{#3}
	}%
}
\NewDocumentCommand\pdf{smmg}{
	{#2%
	\IfBooleanTF{#1}{%
		\BracketsStar{#3}{#4}%
	}{%
		\BracketsNoStar{#3}{#4}%
	}}%
}

\newcommand{\prob}{p\CondBrackets}

\NewDocumentCommand\indicator{sm}{%
	\mathbf{1}%
	\IfValueTF{#1}{%
		\left[#2\right]%
	}{%
		[#2]%
	}%
}

\newcommand{\numberthis}{\addtocounter{equation}{1}\tag{\theequation}}

\newcommand{\DP}{\operatorname{DP}}
\newcommand{\GEM}{\operatorname{GEM}}

\newcommand{\Cat}{\operatorname{Cat}}
\newcommand{\Dir}{\operatorname{Dir}}

\newcommand{\ObsPrior}{H_\mathrm{X}}
\newcommand{\ObsLik}{F_\mathrm{X}}
\newcommand{\ObsPred}{\widetilde{\ObsLik}}

\newcommand{\LabelSetKnown}{\mathcal{Y}}

\newcommand{\LabelPrior}{H_\mathrm{Y}}
\newcommand{\LabelLik}{F_\mathrm{Y}}
\newcommand{\LabelMarg}{\widehat{\LabelLik}}

\newcommand{\LabelPost}{\widehat{\LabelPrior}}
\newcommand{\LabelBase}{L}

\newcommand{\FrameIndices}{\mathcal{F}}
\newcommand{\LabelledIndices}{\mathcal{L}}

\newcommand{\LHS}{\hspace{2em}&\hspace{-2em}}

\def\eg{e.g.~}
\def\ie{i.e.~}

\newcommand\blfootnote[1]{%
  \begingroup
  \renewcommand\thefootnote{}\footnote{#1}%
  \addtocounter{footnote}{-1}%
  \addtocounter{Hfootnote}{-1}%
  \endgroup
}

\begin{document}

\pagestyle{headings}
\mainmatter

\title{From Face Recognition to Models of Identity:~\\
A Bayesian Approach to Learning about~\\
Unknown Identities from Unsupervised Data}

\titlerunning{From Face Recognition to Models of Identity}

\author{
    Daniel Coelho de Castro\inst{1}\fnmsep\thanks{Work done during an internship at Microsoft Research.} \and
    Sebastian Nowozin\inst{2}
}
\authorrunning{D.C. Castro and S. Nowozin}
\institute{
    Imperial College London, UK\\\email{dc315@imperial.ac.uk} \and
    Microsoft Research, Cambridge, UK\\\email{Sebastian.Nowozin@microsoft.com}
}

\maketitle
\blfootnote{Accepted for publication at ECCV 2018.}

\begin{abstract}
Current face recognition systems robustly recognize identities across a wide variety of imaging conditions.
In these systems recognition is performed via classification into known identities obtained from supervised identity annotations.
There are two problems with this current paradigm:
(1) current systems are unable to benefit from unlabelled data which may be available in large quantities; and
(2) current systems equate successful recognition with labelling a given input image.
Humans, on the other hand, regularly perform identification of individuals completely unsupervised, recognising the identity of someone they have seen before even without being able to name that individual.
How can we go beyond the current classification paradigm towards a more human understanding of identities?
%
We propose an integrated Bayesian model that coherently reasons about the observed images, identities, partial knowledge about names, and the situational context of each observation.
While our model achieves good recognition performance against known identities, it can also discover new identities from unsupervised data and learns to associate identities with different contexts depending on which identities tend to be observed together.
In addition, the proposed semi-supervised component is able to handle not only acquaintances, whose names are known, but also unlabelled familiar faces and complete strangers in a unified framework.
%
%
\end{abstract}

\section{Introduction}\label{sec:intro}

For the following discussion, we decompose the usual face identification task into two sub-problems: \emph{recognition} and \emph{tagging}. Here we understand recognition as the unsupervised task of matching an observed face to a cluster of previously seen faces with similar appearance (disregarding variations in pose, illumination etc.), which we refer to as an \emph{identity}. Humans routinely operate at this level of abstraction to recognise familiar faces: even when people's names are not known, we can still tell them apart. Tagging, on the other hand, refers to putting names to faces, \ie associating string literals to known identities.

Humans tend to create an inductive mental model of facial appearance for each person we meet, which we then query at new encounters to be able to recognise them. This is opposed to a transductive approach, attempting to match faces to specific instances from a memorised gallery of past face observations---which is how identification systems are often implemented \cite{Jafri2009}.

An alternative way to represent faces, aligned with our inductive recognition, is via \emph{generative} face models, which explicitly separate latent identity content, tied across all pictures of a same individual, from nuisance factors such as pose, expression and illumination \cite{Ioffe2006,Prince2007,Li2012}. While mostly limited to linear projections from pixel space (or mixtures thereof), the probabilistic framework applied in these works allowed tackling a variety of face recognition tasks, such as closed- and open-set identification, verification and clustering.

\begin{figure}[tb]
	\centering
    \subfloat[][Standard face recognition]
    	{\includegraphics[width=.43\textwidth]{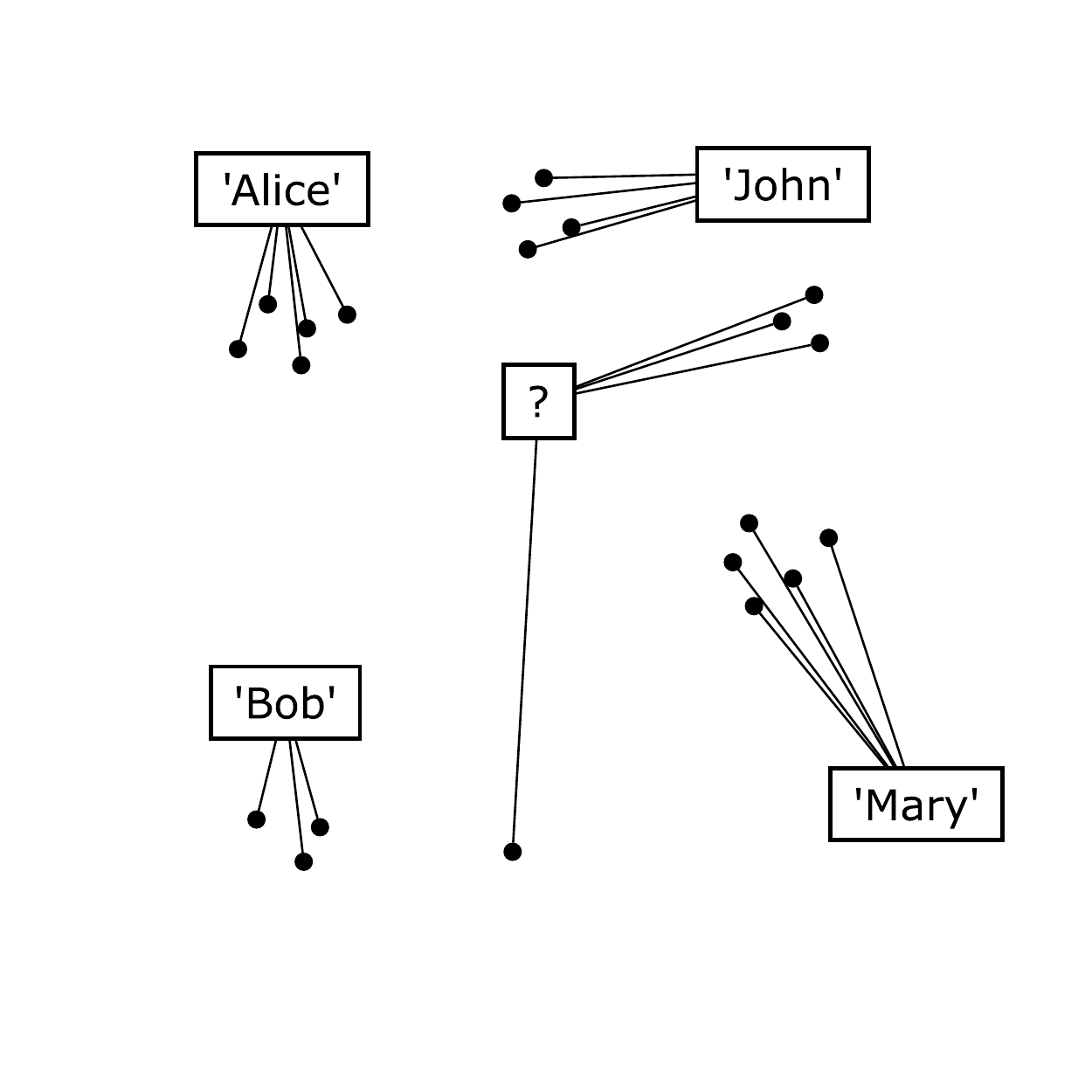}\label{fig:std_recognition_setting}}
    \hspace{2em}
    \subfloat[][Context-aware model of identities]
    	{\includegraphics[width=.43\textwidth]{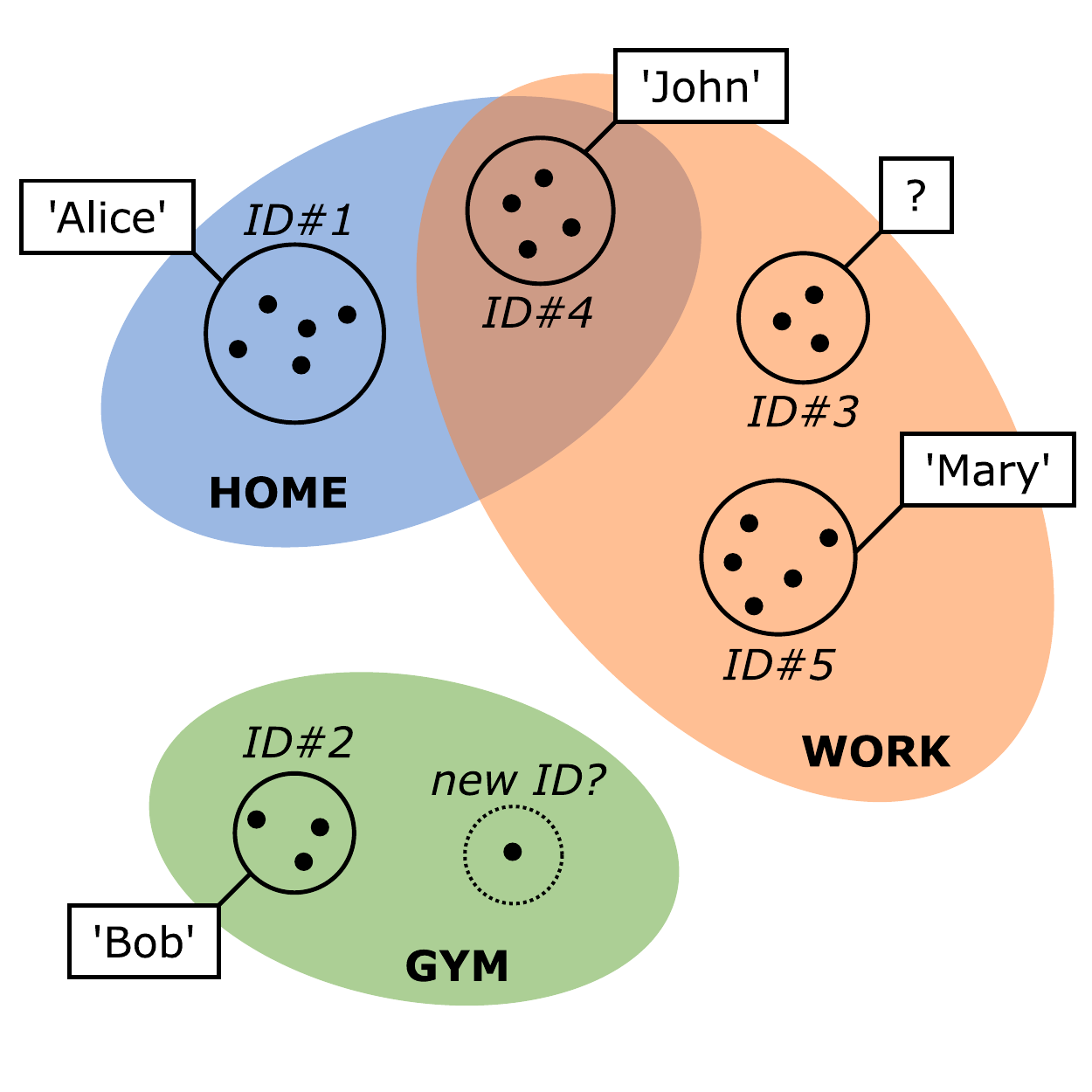}\label{fig:context_aware_setting}}
    \caption{Face recognition settings. Points represent face observations and boxes are name labels.}
\end{figure}

A further important aspect of social interactions is that, as an individual continues to observe faces every day, they encounter some people much more often than others, and the total number of distinct identities ever met tends to increase virtually without bounds. Additionally, we argue that human face recognition does not happen in an isolated environment, but situational contexts (\eg `home', `work', `gym') constitute strong cues for the groups of people a person expects to meet (\cref{fig:context_aware_setting}).

With regards to tagging, in daily life we very rarely obtain named face observations: acquaintances normally introduce themselves only once, and not repeatedly whenever they are in our field of view. In other words, humans are naturally capable of semi-supervised learning, generalising sparse name annotations to all observations of the corresponding individuals, while additionally reconciling naming conflicts due to noise and uncertainty.

In contrast, standard computational face identification is \emph{fully supervised} (see \cref{fig:std_recognition_setting}), relying on vast labelled databases of high-quality images \cite{LearnedMiller2016}. Although many supervised methods achieve astonishing accuracy on challenging benchmarks (\eg \cite{Taigman2014,Schroff2015}) and are successfully employed in practical biometric applications, this setting has arguably limited analogy to human social experience.

Expanding on the generative perspective, we introduce a unified Bayesian model which reflects all the above considerations on identity distributions, context-awareness and labelling (\cref{fig:context_aware_setting}). Our nonparametric identity model effectively represents an unbounded population of identities, while taking contextual co-occurrence relations into account and exploiting modern deep face representations to overcome limitations of previous linear generative models. Our main contributions in this work are twofold:
\begin{enumerate}
	\item We propose an unsupervised face recognition model which can explicitly reason about people it has never seen; and
    \item We attach to it a novel robust label model enabling it to predict names by learning from both named and unnamed faces.
\end{enumerate}

\subsection*{Related Work}\label{sec:related_work}
Other face recognition methods (even those formulated in a Bayesian framework) \cite{Zhang2003,Zhao2006,Choi2010,Tapaswi2012,Le2017}, often limit themselves to point estimates of parameters and predictions, occasionally including ad-hoc confidence metrics. A distinct advantage of our approach is that it is probabilistic end-to-end, and thus naturally provides predictions with principled, quantifiable uncertainties. Moreover, we employ modern Bayesian modelling tools---namely hierarchical nonparametrics---which enable dynamically adapting model complexity while faithfully reflecting the real-world assumptions laid out above.


Secondly, although automatic face tagging is a very common task, each problem setting can impose wildly different assumptions and constraints. Typical application domains involve the annotation of personal photo galleries \cite{Zhang2003,Zhao2006,Anguelov2007,Gallagher2009}, multimedia (\eg TV) \cite{Tapaswi2012,Le2017} or security/surveillance \cite{Jafri2009}. Our work focuses on egocentric human-like face recognition, a setting which seems largely unexplored, as most of the work using first-person footage appears to revolve around other tasks like object and activity recognition, face detection, and tracking \cite{Betancourt2015}. As we explained previously, the dynamic, \emph{online} nature of first-person social experience brings a number of specific modelling challenges for face recognition.


Finally, while there is substantial prior work on using contexts to assist face recognition, we emphasize that much (perhaps most) of it is effectively complementary to our unified framework. Notions of \emph{global} context such as timestamp, geolocation and image background \cite{Torralba2003,Zhao2006,Anguelov2007,Choi2010} can readily be used to inform our current context model (\cref{sec:context_model}). In addition, we can naturally augment the proposed face model (\cref{sec:face_model}) to leverage further \emph{individual} context features, \eg clothing and speech \cite{Zhao2006,Anguelov2007,Tapaswi2012,Le2017}. Integration of these additional factors opens exciting avenues for future research.


\section{A Model of Identities}

\begin{figure}[t]
	\centering
	\tikzset{>=latex}
\tikzstyle{wrap} += [inner sep=1pt]
\tikzstyle{plate} += [inner sep=4pt]
\tikzstyle{const} += [inner sep=3pt, node distance=.5, font=\fontsize{10}{10}\selectfont]
\tikzset{
    diag fill/.style 2 args={fill=#2, path picture={
        \fill[#1, sharp corners] (path picture bounding box.south west) -|
        (path picture bounding box.north east) -- cycle;}}
}
\tikzstyle{partial} = [latent, diag fill={gray!25}{white}]
\tikzstyle{marg} = [latent, dashed]

\colorlet{idtcol}{brown!50!black}
\colorlet{ctxcol}{green!40!black}
\colorlet{labcol}{blue!60!black}
\colorlet{faccol}{red!60!black}

\begin{tikzpicture}[x=36pt, y=48pt]
	\node[obs,draw=faccol,text=faccol,fill=faccol!20]	at ( 0, 0)	(x)		{$\*x_n$};
	\node[latent,draw=idtcol,text=idtcol]	at (-1, 1)	(z)		{$z_n$};
	\node[obs,draw=ctxcol,text=ctxcol,fill=ctxcol!20]	at (-2, 0)	(c)		{$c^*_m$};
    \node[latent,draw=labcol,text=labcol,diag fill={labcol!20}{white}]	at ( 0, 1)	(y)		{$y_n$};
	\node[latent,draw=ctxcol,text=ctxcol]	at (-3, 0)	(omega)	{$\bomega$};
	\node[latent,draw=idtcol,text=idtcol]	at (-2, 1)	(pi)	{$\bpi_c$};
	\node[latent,draw=idtcol,text=idtcol]	at (-3, 1)	(chi)	{$\bpi_0$};
	\node[latent,draw=faccol,text=faccol]	at ( 1, 0)	(theta)	{$\theta^*_i$};
	\node[latent,draw=labcol,text=labcol]	at ( 1, 1)	(l)		{$y^*_i$};
	\node[const,text=idtcol]				at (-3,1.6)	(alpha) {$\alpha_0$};
	\node[const,text=idtcol]				at (-2,1.6)	(epsc)	{$\alpha_c$};
	\node[const,text=ctxcol]				at (-3.8,0)	(gamma)	{$\bgamma$};
	\node[const,text=faccol]				at ( 2, 0)	(hx)	{$\ObsPrior$};
	\node[latent,draw=labcol,text=labcol]	at ( 2, 1)	(hy)	{$\LabelPrior$};
	\node[const,text=labcol]				at (1.8,1.6)(lambda){$\lambda$};
	\node[const,text=labcol]				at (2.2,1.6)(base)	{$\LabelBase$};
    \node[const,text=labcol]				at ( 0,1.6)	(veps)	{$\varepsilon$};
    \node[const,text=ctxcol]				at (-1, 0)	(f)		{$f_n$};
	
	\edge[faccol] {z, theta}	{x};
	\edge[idtcol] {pi}			{z};
	\edge[ctxcol] {c, f}		{z};
	\edge[ctxcol] {omega}		{c};
	\edge[ctxcol] {gamma}		{omega};
	\edge[idtcol] {alpha}		{chi};
	\edge[idtcol] {epsc, chi}	{pi};
	\edge[faccol] {hx}			{theta};
	\edge[labcol] {z, l, veps}	{y};
	\edge[labcol] {hy}			{l};
	\edge[labcol] {lambda, base}{hy};
	
    \plate {N} {(z)(x)(y)(f)}{$N$};
    \plate {M} {(c)}			{$M$};
	\plate {C} {(pi)(epsc)}	{$C$};
	\plate {I} {(theta)(l)}	{$\vphantom{N}\infty$};
    
    \node[above=2pt of C] {context $c$};
    \node[below=2pt of M] {frame $m$};
    \node[below=2pt of N] {observation $n$};
    \node[below=2pt of I] {identity $i$};
\end{tikzpicture}
	\caption{Overview of the proposed generative model, encompassing the \textcolor{ctxcol}{\bf context model}, \textcolor{idtcol}{\bf identity model}, \textcolor{faccol}{\bf face model} and \textcolor{labcol}{\bf label model}. Unfilled nodes represent latent variables, shaded nodes are observed, the half-shaded node is observed only for a subset of the indices and uncircled nodes denote fixed hyperparameters.
    $\bpi_0$ and $(\bpi_c)_{c=1}^C$ are the global and context-wise identity probabilities, $\bomega$ denotes the context probabilities, $(c^*_m)_{m=1}^M$ are the frame-wise context labels, indexed by the frame numbers $(f_n)_{n=1}^N$, $(z_n)_{n=1}^N$ are the latent identity indicators, $(\*x_n)_{n=1}^N$ are the face observations and $(y_n)_{n=1}^N$ are the respective name annotations, $(\theta^*_i)_{i=1}^\infty$ are the parameters of the face model and $(y^*_i)_{i=1}^\infty$ are the identities' name labels. See text for descriptions of the remaining symbols.}
    \label{fig:model_diagram}
\end{figure}
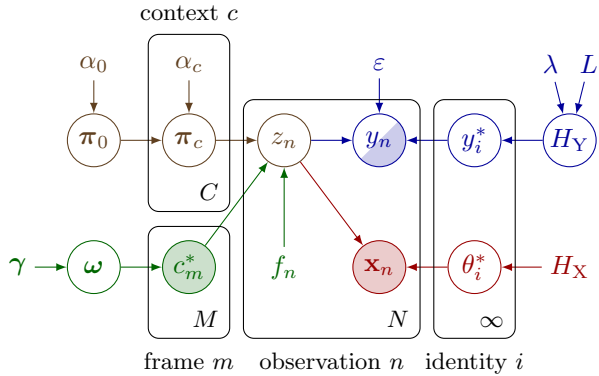

In this section, we describe in isolation each of the building blocks of the proposed approach to facial identity recognition: the context model, the identity model and the face model. We assume data is collected in the form of camera \emph{frames} (either photographs or a video stills), numbered $1$ to $M$, and faces are cropped with some face detection system and grouped by frame number indicators, $f_n \in \{1,\dots,M\}$. The diagram in \cref{fig:model_diagram} illustrates the full proposed graphical model, including the label model detailed in \cref{sec:label_model}.

\subsection{Context Model}\label{sec:context_model}


In our identity recognition scenario, we imagine the user moving between contexts throughout the day (\eg home--work--gym...). Since humans naturally use situational context as a strong prior on the groups of people we expect to encounter in each situation, we incorporate context-awareness in our model of identities to mimic human-like face recognition.

The context model we propose involves a categorical variable $c_n \in {\{1,\dots,C\}}$ for each observation, where $C$ is some fixed number of distinct contexts.\footnote{See footnote \labelcref{foot:unbounded_contexts}.} Crucially, we assume that all observations in frame $m$, $\FrameIndices_m = {\{n : f_n = m\}}$, share the same context, $c^*_m$ (\ie $\forall n, c_n = c^*_{f_n}$).

We define the identity indicators to be independent given the context of the corresponding frames (see \cref{sec:id_model}, below). However, since the contexts are tied by frame, marginalising over the contexts captures identity co-occurrence relations. In turn, these allow the model to make more confident predictions about people who tend to be seen together in the same environment.

This formalisation of contexts as discrete semantic labels is closely related to the place recognition model in \cite{Torralba2003}, used there to disambiguate predictions for object detection. It has also been demonstrated that explicit incorporation of a context variable can greatly improve clustering with mixture models \cite{Perdikis2015}.


Finally, we assume the context indicators $c^*_m$ are independently distributed according to probabilities $\bomega$, which themselves follow a Dirichlet prior:
\begin{align}
	\bomega					&\sim \Dir(\bgamma) \\
	c^*_m \cond \bomega		&\sim \Cat(\bomega) \,, & m &= 1,\dots,M \,,
\end{align}
where $M$ is the total number of frames. In our simulation experiments, we use a symmetric Dirichlet prior, setting $\bgamma = (\gamma_0 / C, \dots, \gamma_0 / C)$.

\subsection{Identity Model}\label{sec:id_model}

In the daily-life scenario described in \cref{sec:intro}, an increasing number of unique identities will tend to appear as more faces are observed. This number is expected to grow much more slowly than the number of observations, and can be considered unbounded in practice (we do not expect a user to run out of new people to meet). Moreover, we can expect some people to be encountered much more often than others. Since a Dirichlet process (DP) \cite{Ferguson1973} displays properties that mirror all of the above phenomena \cite{Teh2010}, it is a sound choice for modelling the distribution of identities.

Furthermore, the assumption that all people can potentially be encountered in any context, but with different probabilities, is perfectly captured by a hierarchical Dirichlet process (HDP) \cite{Teh2006}. Making use of the context model, we define one DP \emph{per context} $c$, each with concentration parameter $\alpha_c$ and sharing the same \emph{global} DP as a base measure.\footnote{\label{foot:unbounded_contexts}One could further allow an unbounded number of latent contexts by incorporating a nonparametric context distribution, resulting in a structure akin to the nested DP \cite{Rodriguez2008,Blei2010} or the dual DP described in \cite{Wang2009b}. See \cref{app:random_measure} for details.} This hierarchical construction thus produces context-specific distributions over a common set of identities.

We consider that each of the $N$ face detections is associated to a latent identity indicator variable, $z_n$. We can write the generative process as
\begin{align}
	\bpi_0 				&\sim \GEM(\alpha_0) \\
	\bpi_c \cond \bpi_0	&\sim \DP(\alpha_c, \bpi_0) \,,					& c &= 1,\dots,C \\
	z_n \cond f_n = m, \*c^*, (\bpi_c)_c	&\sim \Cat(\bpi_{c^*_m})\,,	& n &= 1,\dots,N \,,
\end{align}
where $\GEM(\alpha_0)$ is the DP stick-breaking distribution, ${\pi_{0i} = \beta_i \prod_{j=1}^{i-1} (1-\beta_j)}$, with ${\beta_i \sim \operatorname{Beta}(1,\alpha_0)}$ and $i=1,\dots,\infty$. Here, $\bpi_0$ is the global identity distribution and $(\bpi_c)_{c=1}^C$ are the context-specific identity distributions.

Although the full generative model involves infinite-dimensional objects, DP-based models present simple finite-dimensional marginals. In particular, the posterior predictive probability of encountering a known identity $i$ is
\begin{equation}\label{eq:prob_known}
	\prob{z_{N+1} = i}{c_{N+1}=c, \*z, \*c^*, \bpi_0}
		= \frac{\alpha_c \pi_{0i} + N_{ci}}{\alpha_c + N_{c \cdot}} \,,
\end{equation}
where $N_{ci}$ is the number of observations assigned to context $c$ and identity $i$ and $N_{c\cdot}$ is the total number of observations in context $c$.

Finally, such a nonparametric model is well suited for an open-set identification task, as it can elegantly estimate the prior probability of encountering an unknown identity:
\begin{equation}\label{eq:prob_unknown}
	\prob{z_{N+1} = I+1}{c_{N+1}=c, \*z, \*c^*, \bpi_0}
		= \frac{\alpha_c \pi_0'}{\alpha_c + N_{c \cdot}} \,,
\end{equation}
where $I$ is the current number of distinct known identities and ${\pi_0' = \sum_{i=I+1}^\infty \pi_{0i}}$ denotes the global probability of sampling a new identity.

\subsection{Face Model}\label{sec:face_model}

In face recognition applications, it is typically more convenient and meaningful to extract a compact representation of face features than to work directly in a high-dimensional pixel space.

We assume that the observed features of the $n$\textsuperscript{th} face, $\*x_n$, arise from a parametric family of distributions, $\ObsLik$. The parameters of this distribution, $\theta^*_i$, drawn from a prior, $\ObsPrior$, are unique for each identity and are shared across all face feature observations of the same person:
\begin{align}
	\theta^*_i					&\sim \ObsPrior \,,					& i &= 1,\dots,\infty \\
	\*x_n \cond z_n, \btheta^*	&\sim \ObsLik(\theta^*_{z_n}) \,,	& n &= 1,\dots,N \,.
\end{align}
As a consequence, the marginal distribution of faces is given by a \emph{mixture model}: $\prob{\*x_n}{c_n=c, \btheta^*, \bpi_c} = \sum_{i=1}^\infty \pi_{ci} \ObsLik(\*x_n \cond \theta^*_i)$.

In the experiments reported in this paper, we used the 128-dimensional embeddings produced by OpenFace, a publicly available, state-of-the-art neural network for face recognition \cite{Amos2016}, implementing FaceNet's architecture and methodology \cite{Schroff2015}. In practice, this could easily be swapped for other face embeddings (\eg DeepFace \cite{Taigman2014}) without affecting the remainder of the model. We chose isotropic Gaussian mixture components for the face features ($\ObsLik$), with an empirical Gaussian--inverse gamma prior for their means and variances ($\ObsPrior$).




\section{Robust Semi-Supervised Label Model}\label{sec:label_model}


We expect to work with only a small number of labelled observations manually provided by the user. Since the final goal is to identify any observed face, our probabilistic model needs to incorporate a semi-supervised aspect, generalising the sparse given labels to unlabelled instances. Throughout this section, the terms `identity' and `cluster' will be used interchangeably.

\begin{figure}[tb]
	\centering
    \includegraphics[width=.48\textwidth]{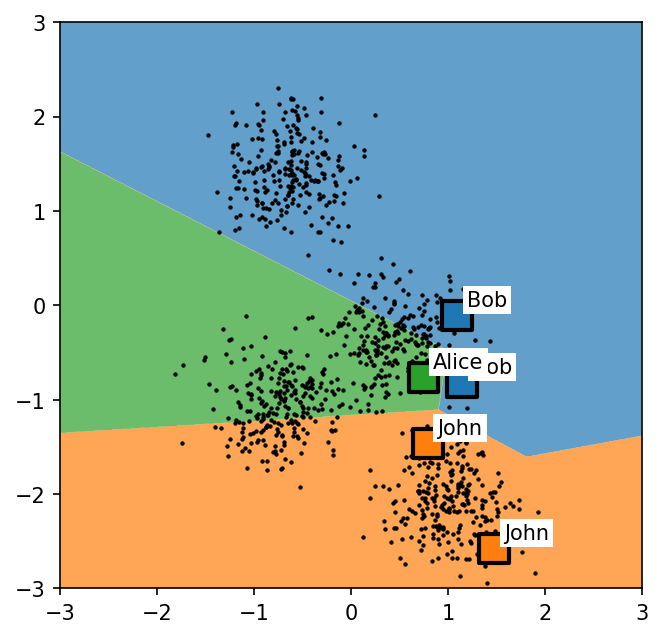}
    \hfill
    \includegraphics[width=.48\textwidth]{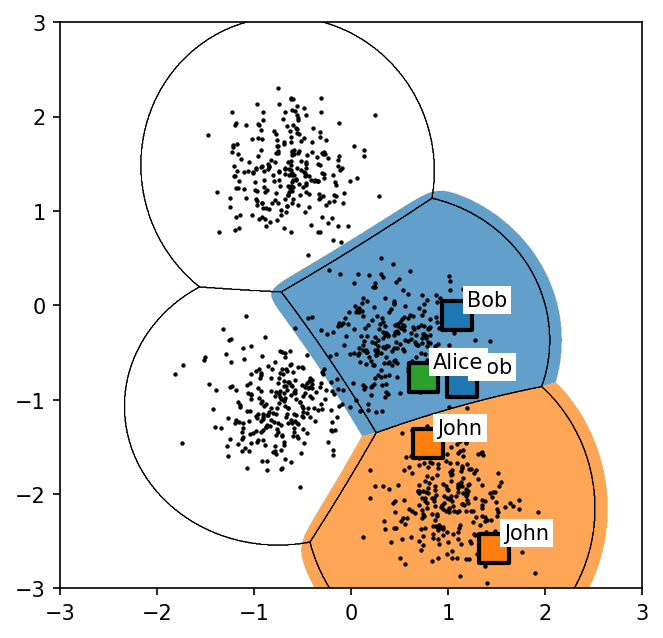}
    \caption{Hard label predictions of the proposed semi-supervised label model (right) and nearest-neighbour classification (left). Points represent unlabelled face observations, squares are labelled and the black contours on the right show identity boundaries. The proposed label model produces more \emph{natural} boundaries, assigning the `unknown' label (white) to unlabelled clusters and regions distant from any observed cluster, while also accommodating label noise (`Bob' $\to$ `Alice') without the spurious boundaries introduced by NN.}
	\label{fig:labels}
\end{figure}

One of the cornerstones of semi-supervised learning (SSL) is the premise that clustered items tend to belong to the same class \cite[\S 1.2.2]{Chapelle2006}. Building on this \emph{cluster assumption}, mixture models, such as ours, have been successfully applied to SSL tasks \cite{Bouveyron2009}. We illustrate in \cref{fig:labels} our proposed label model detailed below, comparing it qualitatively to nearest-neighbour classification on a toy example.

With the motivation above, we attach a label variable (a \emph{name}) to each cluster (identity), here denoted $y^*_i$. This notation suggests that there is a single true label ${\tilde{y}_n = y^*_{z_n}}$ for each observation $n$, analogously to the observation parameters: ${\theta_n = \theta^*_{z_n}}$. Finally, the observed labels, $y_n$, are potentially corrupted through some noise process, $\LabelLik$. Let $\LabelledIndices$ denote the set of indices of the labelled data. The complete generative process is presented below:
\begin{align}
	\LabelPrior				&\sim \DP(\lambda, \LabelBase) \\
	y^*_i \cond \LabelPrior	&\sim \LabelPrior \,,					& i &= 1,\dots,\infty \\
	y_n \cond z_n, \*y^*, \LabelPrior	&\sim \LabelLik(y^*_{z_n}; \LabelPrior) \,,			& n &\in \LabelledIndices \,.
\end{align}

As mentioned previously, a related model for mixture model-based SSL with noisy labels was proposed in \cite{Bouveyron2009}. Instead of considering an explicit noise model for the class labels, the authors of that work model directly the conditional label distribution for each cluster. Our setting here is more general: we assume not only an unbounded number of clusters, but also of possible labels.

\subsection{Label Prior}
We assume that the number of distinct labels will tend to increase without bounds as more data is observed. Therefore, we adopt a further nonparametric prior on the cluster-wide labels:
\begin{equation}
	\LabelPrior \sim \DP(\lambda, \LabelBase) \,,
\end{equation}
where $\LabelBase$ is some base probability distribution over the countable but unbounded label space (\eg strings).\footnote{One could instead consider a Pitman--Yor process if power-law behaviour seems more appropriate than the DP's exponential tails \cite{Pitman1997}.}
We briefly discuss the choice of $\LabelBase$ further below.

All concrete knowledge we have about the random label prior $\LabelPrior$ comes from the set of observed labels, $\*y_\LabelledIndices$. Crucially, if we marginalise out $\LabelPrior$, the predictive label distribution is simply \cite{Teh2010}
\begin{equation}\label{eq:label_posterior}
	y^*_{I+1} \cond \*y^* \sim \frac{1}{\lambda + I} \biggl( \lambda \LabelBase + \sum_{\ell \in \LabelSetKnown} J_\ell \delta_\ell \biggr) \,,
\end{equation}
which we will denote $\LabelPost(y^*_{I+1} \cond \*y^*)$. Here, $\LabelSetKnown$ is the set of distinct known labels among $\*y_\LabelledIndices$ and $J_\ell = |\{i : y^*_i = \ell\}|$, the number of components with label $\ell$ (note that $\sum_\ell J_\ell = I$).






In addition to allowing multiple clusters to have repeated labels, this formulation allows us to reason about \emph{unseen} labels. For instance, some of the learned clusters may have no labelled training points assigned to them, and the true (unobserved) labels of those clusters may never have been encountered among the training labels. Another situation in which unseen labels come into play is with points away from any clusters, for which the identity model would allocate a new cluster with high probability. In both cases, this model gives us a principled estimate of the probability of assigning a special `unknown' label.


The base measure $\LabelBase$ may be defined over a rudimentary language model. For this work, we adopted a geometric/negative binomial model for the string length $|\ell|$, with characters drawn uniformly from an alphabet of size $K$:
\begin{equation}
	\LabelBase_{\phi, K}(\ell) = \operatorname{Geom}(|\ell|; \tfrac{1}{\phi}) \operatorname{Unif}(\ell; K^{|\ell|})
    	= \frac{1}{\phi-1} \left( \frac{\phi - 1}{\phi K} \right)^{|\ell|} \,,
\end{equation}
where $\phi$ is the expected string length.

\subsection{Label Likelihood}
In the simplest case, we could consider $\LabelLik(\cdot) = \delta_\cdot$, \ie noiseless labels. Although straightforward to interpret and implement, this could make inference highly unstable whenever there would be conflicting labels for an identity. Moreover, in our application, the labels would be provided provided by a human user who may not have perfect knowledge of the target person's true name or its spelling, for example.

Therefore, we incorporate a label noise model, which can gracefully handle conflicts and mislabelling. We assume observed labels are noisy completely at random (NCAR) \cite[\S II-C]{Frenay2014}, with a fixed error rate $\varepsilon$:%
\footnote{The `true' label likelihood $\LabelLik(\ell \cond y^*_i; \LabelPrior)$ is random due to its dependence on the unobserved prior $\LabelPrior$. We thus define $\LabelMarg$ as its posterior expectation given the known identity labels $\*y^*$. See \cref{app:label_lik} for details.}
\begin{equation}\label{eq:label_marginal}
	\LabelMarg(\ell \cond y^*_i; \*y^*) = \begin{cases}
		1-\varepsilon \,,	& \ell = y^*_i \\
		\varepsilon \frac{\LabelPost(\ell \cond \*y^*)}{1-\LabelPost(y^*_i \cond \*y^*)} \,,	& \ell \neq y^*_i
	\end{cases} \,.
\end{equation}
Intuitively, an observed label, $y_n$, agrees with its identity's assigned label, $y^*_{z_n}$, with probability $1-\varepsilon$. Otherwise, it is assumed to come from a modified label distribution, in which we restrict and renormalise $\LabelPost$ to exclude $y^*_{z_n}$. Here we use $\LabelPost$ in the error distribution instead of $\LabelBase$ to reflect that a user is likely to mistake a person's name for another known name, rather than for a completely random string.

\subsection{Label Prediction}
For label prediction, we are only concerned with the true, noiseless labels, $\tilde{y}_n$. The predictive distribution for a single new sample is given by
\begin{equation}
\begin{split}
	\LHS \prob{\tilde{y}_{N+1} = \ell}{\*x_{N+1}, \*z, \*c^*, \*y^*, \btheta^*, \bpi_0} \\
    	&= \sum_{i \leq I : y^*_i = \ell} \prob{z_{N+1} = i}{\*x_{N+1}, \*z, \*c^*, \btheta^*, \bpi_0} \\
    	&\qquad + \LabelPost(y^*_{I+1} = \ell \cond \*y^*) \, \prob{z_{N+1} = I+1}{\*x_{N+1}, \*z, \*c^*, \btheta^*, \bpi_0} \,.
\end{split}
\end{equation}
The sum in the first term is the probability of the sample being assigned to any of the existing identities that have label $\ell$, while the last term is the probability of instantiating a new identity with that label.



\section{Evaluation}

One of the main strengths of the proposed model is that it creates a single rich representation of the known world, which can then be queried from various angles to obtain distinct insights. In this spirit, we designed three experimental setups to assess different properties of the model: detecting whether a person has been seen before (outlier detection), recognising faces as different identities in a sequence of frames (clustering, unsupervised) and correctly naming observed faces by generalising sparse user annotations (semi-supervised learning).

In all experiments, we used celebrity photographs from the Labelled Faces in the Wild (LFW) database \cite{Huang2007}.\footnote{Available at: \url{http://vis-www.cs.umass.edu/lfw/}} We have implemented inference via Gibbs Markov chain Monte Carlo (MCMC) sampling, whose conditional distributions can be found in \cref{app:gibbs}, and we run multiple chains with randomised initial conditions to better estimate the variability in the posterior distribution. For all metrics evaluated on our model, we report the estimated 95\% highest posterior density (HPD) credible intervals over pooled samples from 8 independent Gibbs chains, unless stated otherwise.

\subsection{Experiment 1: Unknown Person Detection}\label{sec:eval_unknown}

In our first set of experiments, we study the model's ability to determine whether or not a person has been seen before. This key feature of the proposed model is evaluated based on the probability of an observed face not corresponding to any of the known identities, as given by \cref{eq:prob_unknown}. In order to evaluate purely the detection of unrecognised faces, we constrained the model to a single context ($C=1$) and set aside the label model ($\LabelledIndices = \emptyset$).

This task is closely related to outlier/anomaly detection. In particular, our proposed approach mirrors one of its common formulations, involving a mixture of a `normal' distribution, typically fitted to some training data, and a flatter `anomalous' distribution\footnote{The predictive distribution of $\*x_n$ for new identities is a wide Student's $t$.} \cite[\S 7.1.3]{Chandola2009}.

We selected the 19 celebrities with at least 40 pictures available in LFW and randomly split them in two groups: 10 known and 9 unknown people. We used 27 images of each of the \emph{known} people as training data and a disjoint test set of 13 images of each of the \emph{known} and \emph{unknown} people. We therefore have a binary classification setting with well-balanced classes at test time. Here, we ran our Gibbs sampler for 500 steps, discarding the first 100 burn-in iterations and thinning by a factor of 10, resulting in 320 pooled samples.

\begin{figure}[t]
	\centering
    \subfloat[][Association matrix, counting agreements in the MAP identity predictions (including \emph{unknown}). Ticks delimit ground-truth identities.]
    	{\includegraphics[scale=.44]{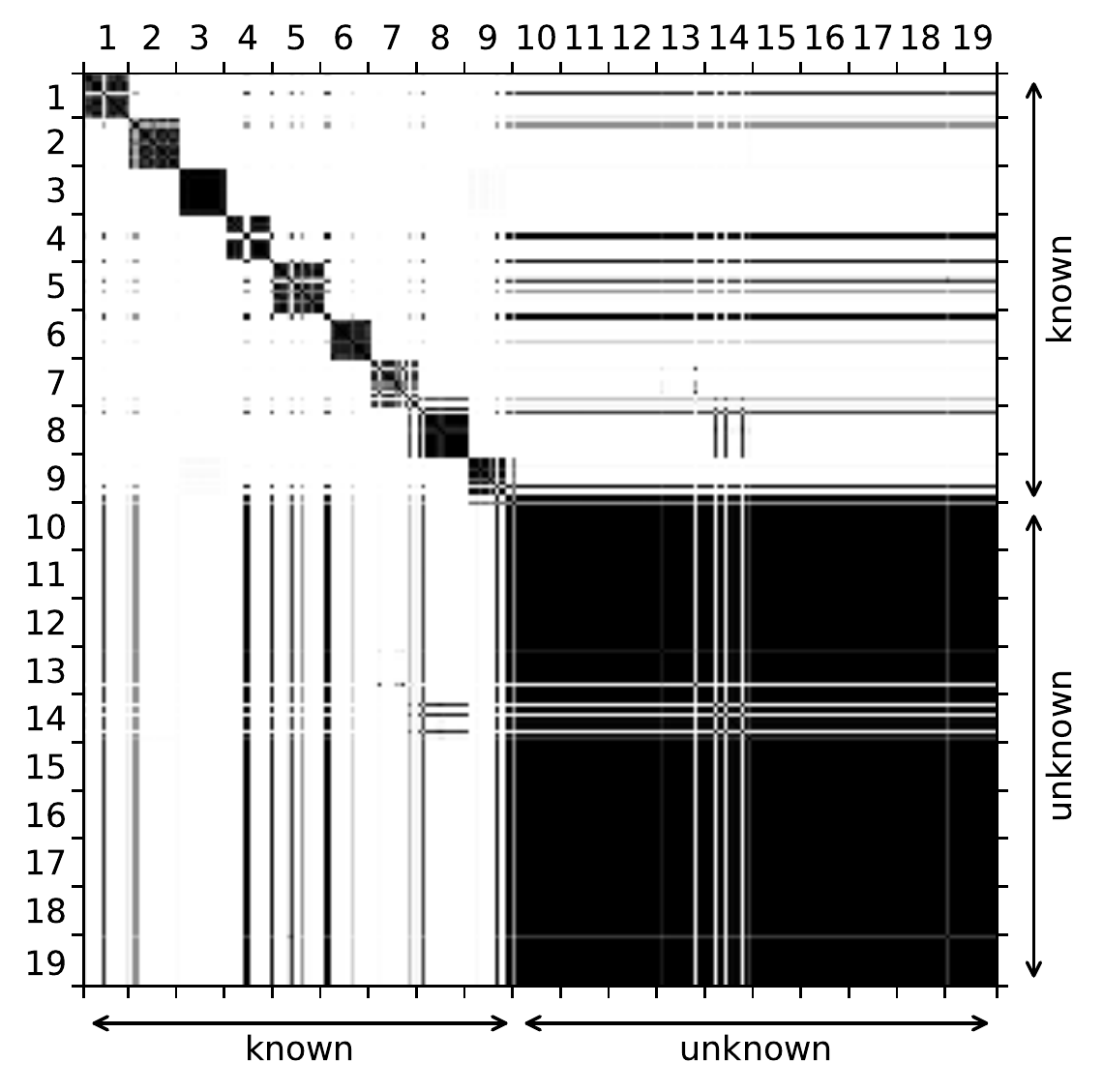}\label{fig:assoc_matrix}}
    \hfill
    \subfloat[][ROC analysis for unknown person detection compared with baselines. AUC is shown with median and 50\% and 95\% HPD  intervals.]
    	{\includegraphics[scale=.44]{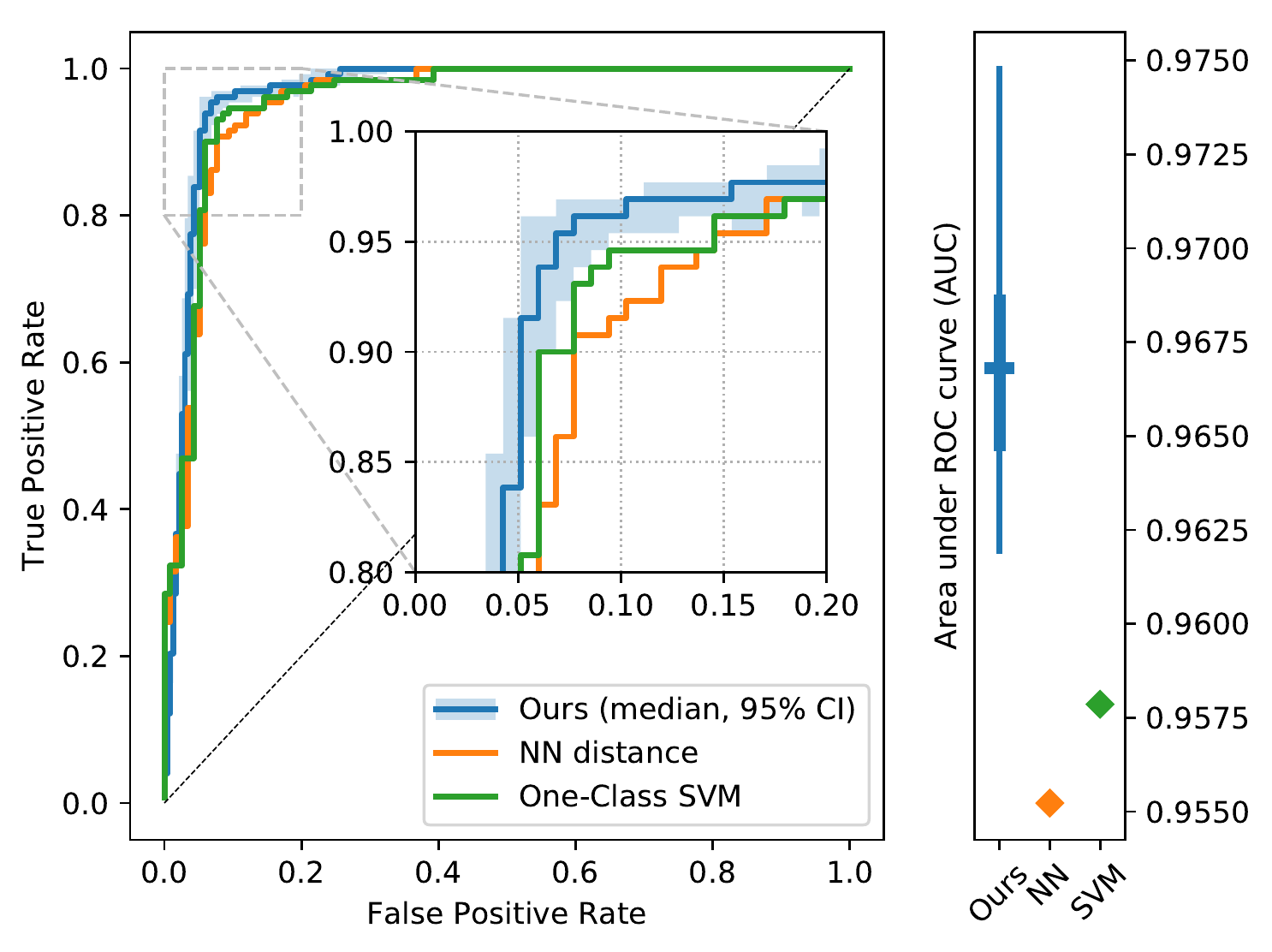}\label{fig:unknown_roc}}
    \caption{Results of the unknown person detection experiment on test images}
\end{figure}

In \cref{fig:assoc_matrix}, we visualise the agreements between maximum \textit{a posteriori} (MAP) identity predictions for test images:
\begin{equation}
	\hat{z}_n = \arg\max_i \prob{z_n=i}{\*x_n, \*z, \*c^*, \bpi_0, \btheta^*} \,,
\end{equation}
where $i$ ranges from $1$ to $I+1$, the latter indicating an \emph{unknown} identity, absent from the training set, and $n$ indexes the test instances. Despite occasional ambiguous cases, the proposed model seems able to consistently group together all unknown faces, while successfully distinguishing between known identities.

As a simple baseline detector for comparison, we consider a threshold on the distance to the nearest neighbour (NN) in the face feature space \cite[\S 5.1]{Chandola2009}. We also evaluate the decision function of a one-class SVM \cite{Scholkopf2001}, using an RBF kernel with $\gamma = 10$, chosen via leave-one-person-out cross-validation on the training set (roughly equivalent to thresholding the training data's kernel density estimate with bandwidth $1/\sqrt{2\gamma} \approx 0.22$). We compare the effectiveness of both detection approaches using ROC curve analysis.

\Cref{fig:unknown_roc} shows that, while all methods are highly effective at detecting unknown faces, scoring $95\%+$ AUC, ours consistently outperforms, by a small margin, both the NN baseline and the purpose-designed one-class SVM. Taking the MAP prediction, our model achieves $[92.3\%, 94.3\%]$ detection accuracy.

\subsection{Experiment 2: Identity Discovery}

We then investigate the clustering properties of the model in a purely unsupervised setting, when only context is provided. We evaluate the consistency of the estimated partitions of images into identities with the ground truth in terms of the adjusted Rand index \cite{Rand1971,Hubert1985}.

\begin{figure}[t]
	\centering
    \includegraphics[width=.9\linewidth]{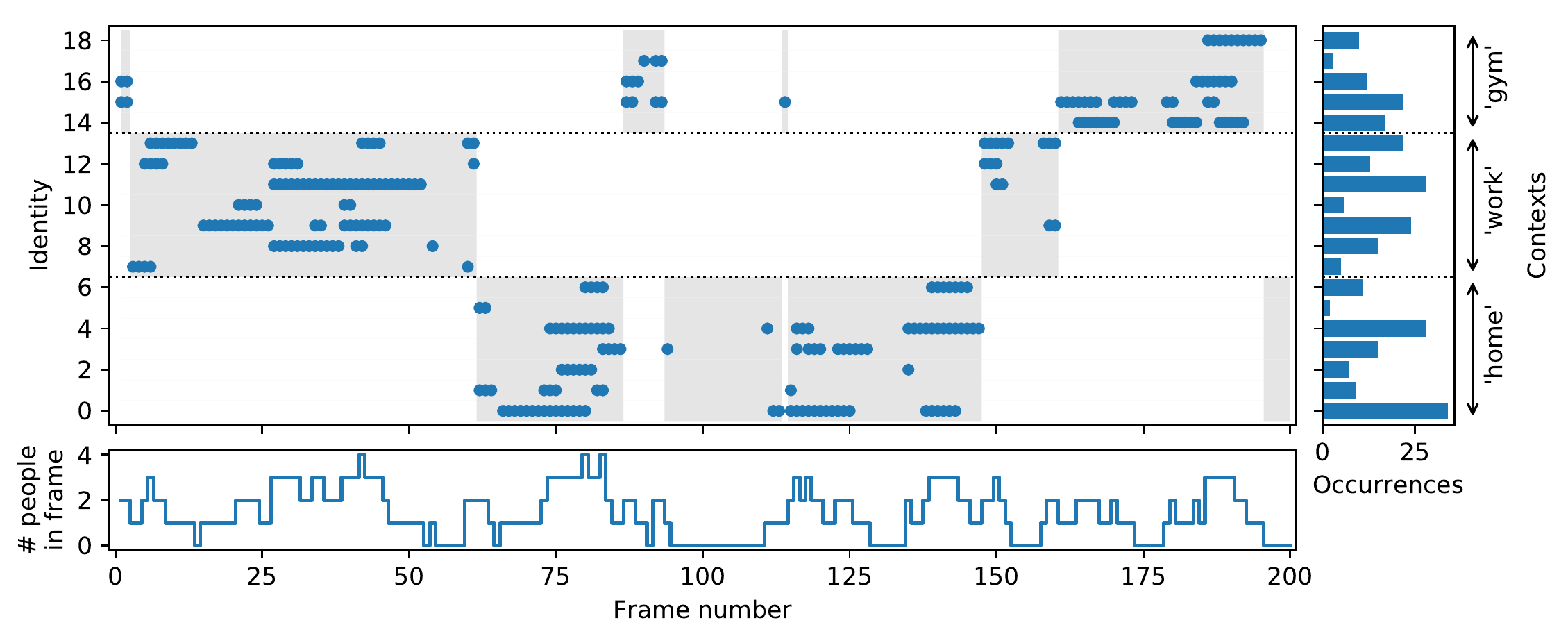}
    \caption{The simulation used in Experiment 2, showing identities coming in and out of the camera frame. Identities are shown grouped by their context (far right), and shading indicates identities present in the user's current context.}
    \label{fig:frame_simulation}
\end{figure}

Using simulations, besides having an endless source of data with ground-truth context and identity labels, we have full control over several important aspects of experimental setup, such as sequence lengths, rates of encounters, numbers of distinct contexts and people and amount of provided labels. Below we describe the simulation algorithm used in our experiments and illustrated in \cref{fig:frame_simulation}.

In our experiments we aim to simulate two important aspects of real-world identity recognition settings:
1. \emph{Context}: knowing the context (\eg location or time) makes it more likely for us to observe a particular subset of people; and
2. \emph{Temporal consistency}: identities will not appear and disappear at random but instead be present for a longer duration.

To reproduce contexts, we simulate a single session of a user meeting new people. To this end we first create a number of fixed contexts and then assign identities uniformly at random to each context. For these experiments, we defined three contexts: `home', `work' and `gym'. At any time, the user knows its own context and over time transitions between contexts. Independently at each frame, the user may switch context with a small probability.

To simulate temporal consistency, each person in the current context enters and leaves the camera frame as an independent binary Markov chain. As shown in \cref{fig:frame_simulation} this naturally produces grouped observations. The image that is observed for each `detected' face is sampled from the person's pictures available in the database. We sample these images without replacement and in cycles, to avoid observing the same image consecutively.

For this set of experiments, we consider three practical scenarios:
\begin{itemize}
	\item \emph{Online:} data is processed on a frame-by-frame basis, \ie we extend the training set after each frame and run the Gibbs sampler for 10 full iterations
    \item \emph{Batch:} same as above, but enqueue data for 20 frames before extending the training set and updating the model for 200 steps
    \item \emph{Offline:} assume entire sequence is available at once and iterate for 1000 steps
\end{itemize}

In the interest of fairness, the number of steps for each protocol was selected to give them roughly the same overall computation budget (ca.~200\,000 frame-wise steps). In addition, we also study the impact on recognition performance of disabling the context model, by setting $C=1$ and $c^*_m=1, \forall m$.

\begin{figure}[t]
	\centering
    \subfloat[][With contexts]
    	{\includegraphics[width=.49\linewidth]{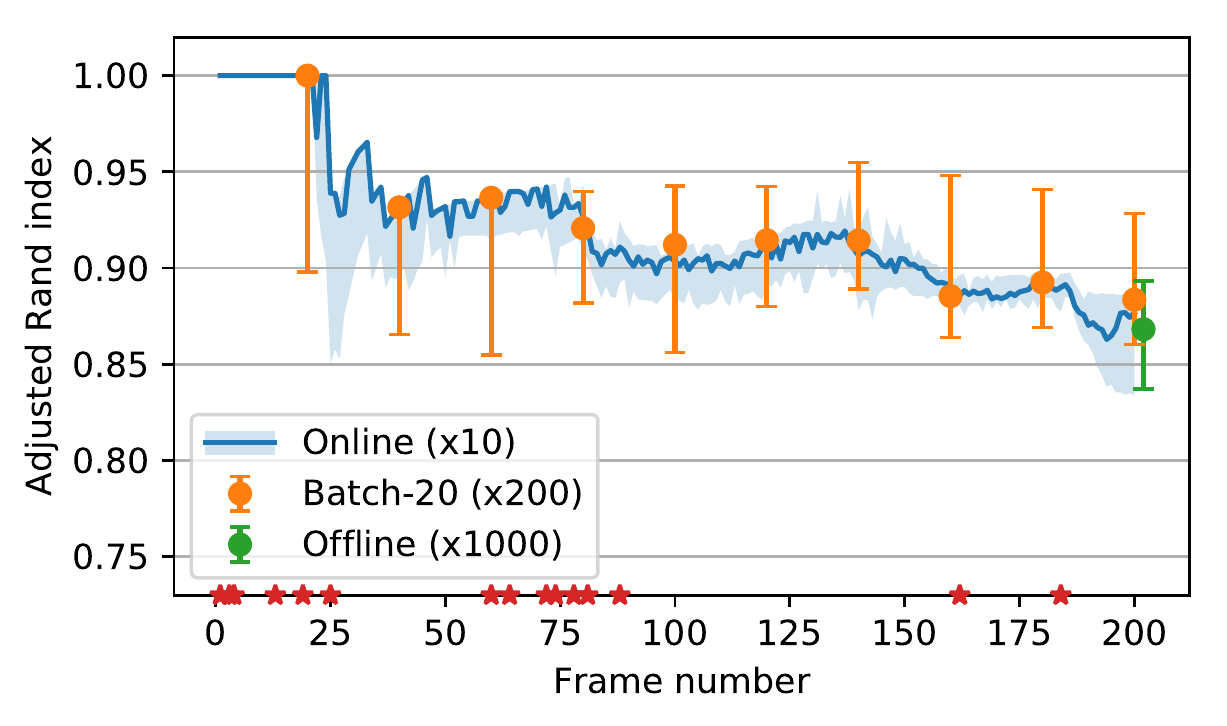}\label{fig:fixed_contexts}}
    \hfill
    \subfloat[][Without contexts]
    	{\includegraphics[width=.49\linewidth]{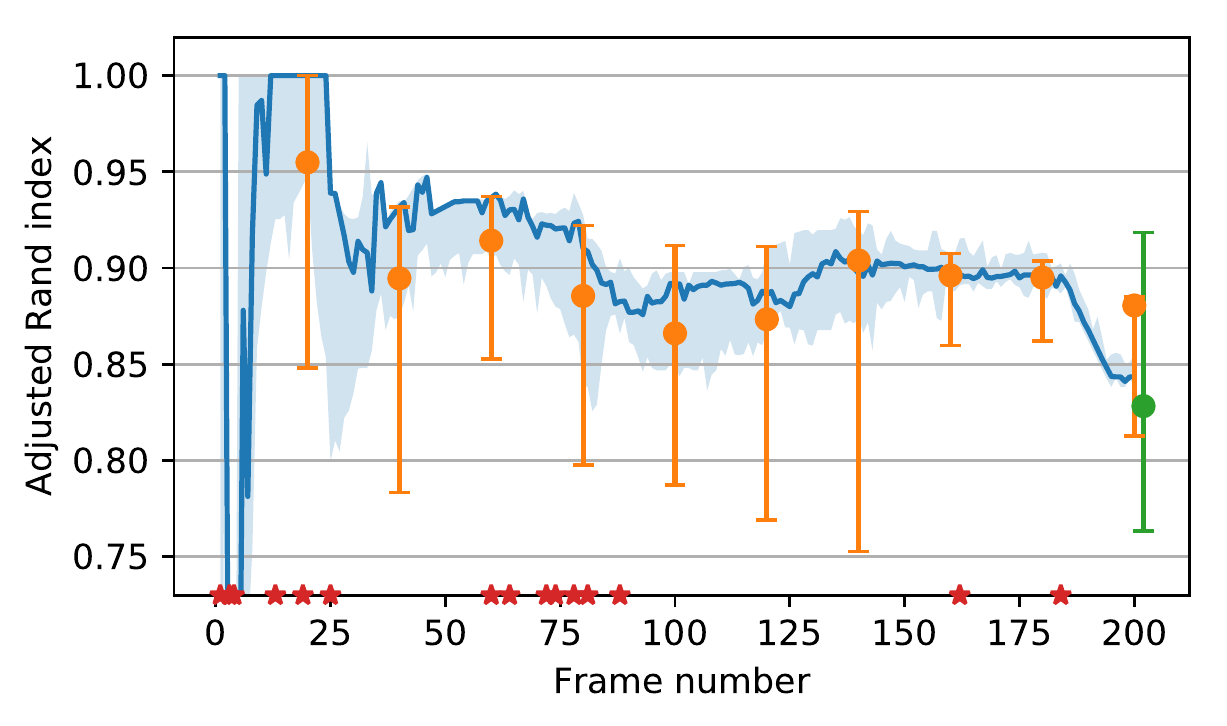}\label{fig:no_contexts}}
    \caption{Identity clustering consistency. Markers on the horizontal axis (\protect\tikz{\protect\node[star,fill=red!80!black,star point ratio=2.25,inner sep=1pt] {};}) indicate when new people are met for the first time.}
    \label{fig:discovery_results}
\end{figure}

We show the results of this experiment in \cref{fig:discovery_results}. Clearly it is expected that, as more identities are met over time, the problem grows more challenging and clustering performance tends to decrease. Another general observation is that online processing produced much lower variance than batch or offline in both cases. The incremental availability of training data therefore seems to lead to more coherent states of the model.

Now, comparing \cref{fig:fixed_contexts,fig:no_contexts}, it is evident that context-awareness not only reduces variance but also shows marginal improvements over the context-oblivious variant. Thus, without hurting recognition performance, the addition of a context model enables the \emph{prediction} of context at test time, which may be useful for downstream user-experience systems.


\subsection{Experiment 3: Semi-Supervised Labelling}

In our final set of experiments, we aimed to validate the application of the proposed label model for semi-supervised learning with sparse labels. 

In the context of face identification, we may define three groups of people:
\begin{itemize}
	\item \emph{Acquainted:} known identity with known name
    \item \emph{Familiar:} known identity with unknown name
    \item \emph{Stranger:} unknown identity
\end{itemize}
We thus selected the 34 LFW celebrities with more than 30 pictures, and split them roughly equally in these three categories at random. From the \emph{acquainted} and \emph{familiar} groups, we randomly picked 15 of their images for training and 15 for testing, and we used 15 pictures of each \emph{stranger} at test time only. We evaluated the label prediction accuracy as we varied the number of labelled training images provided for each acquaintance, from 1 to 15.

For baseline comparison, we evaluate nearest-neighbour classification (NN) and label propagation (LP) \cite{Zhu2002}, a similarity graph-based semi-supervised algorithm. We computed the LP edge weights with the same kernel as the SVM in \cref{sec:eval_unknown}. Recall that the face embedding network was trained with a triplet loss to explicitly optimise Euclidean distances for classification \cite{Amos2016}. As both NN and LP are distance-based, they are therefore expected to hold an advantage over our model for classifying labelled identities.


\begin{figure}[t]
	\centering
    \subfloat[][Acquaintances]
    	{\includegraphics[width=.49\linewidth]{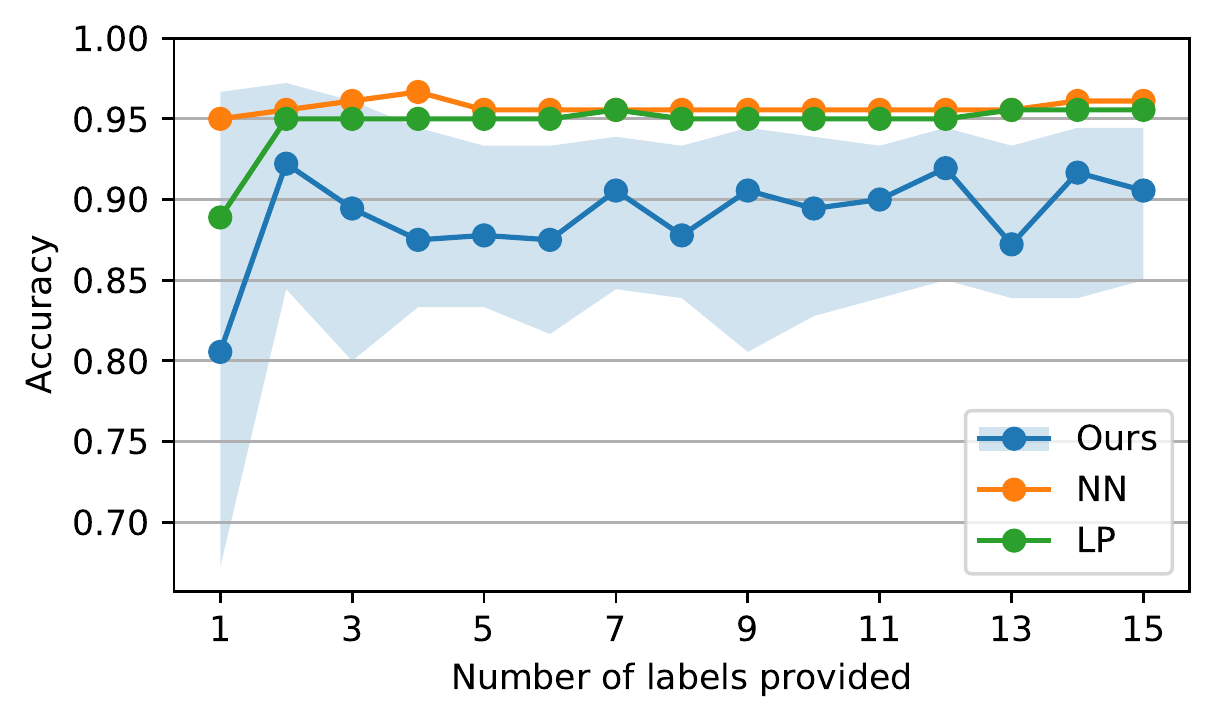}\label{fig:ssl_acqu}}
    \hfill
    \subfloat[][Familiar and strangers]
    	{\includegraphics[width=.49\linewidth]{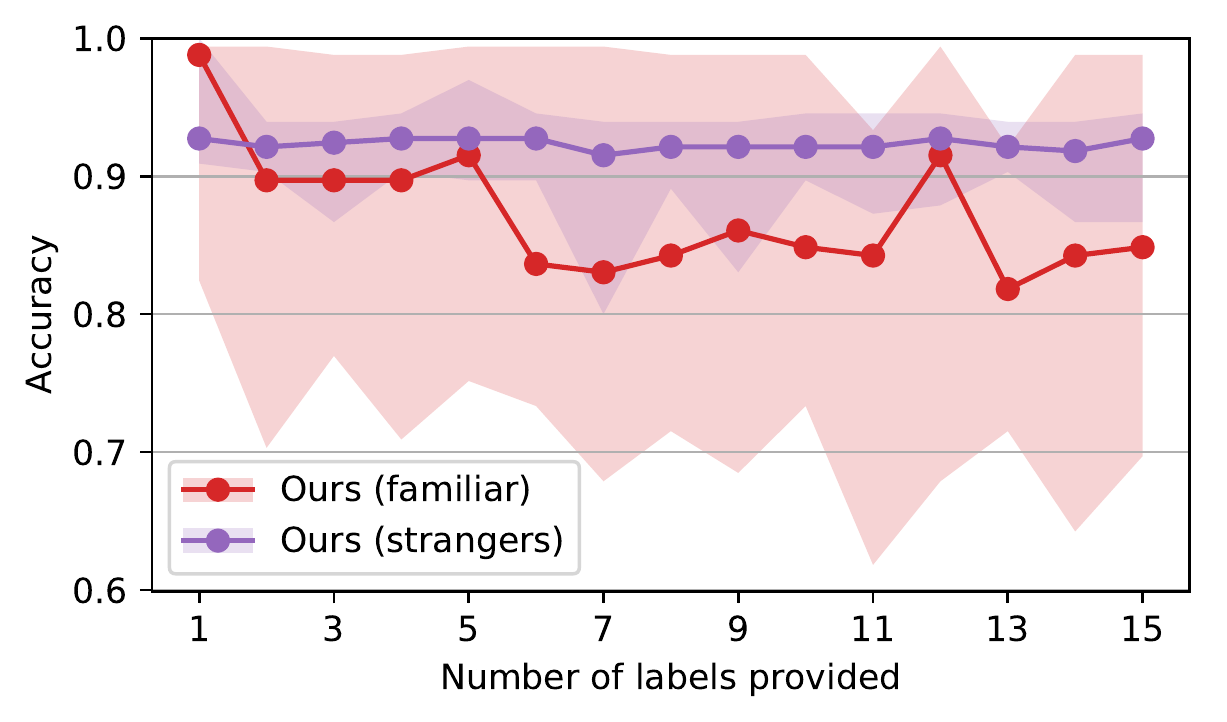}\label{fig:ssl_fami_stra}}
    \caption{Label prediction accuracy. Note that NN and LP effectively have null accuracy for the \emph{familiar} and \emph{strangers} groups, as they cannot predict `unknown'.}
    \label{fig:ssl_results}
\end{figure}


\Cref{fig:ssl_acqu} shows the label prediction results for the labelled identities (acquaintances).
In this setting, NN and LP performed nearly identically and slightly better than ours, likely due to the favourable embedding structure.
Moreover, all methods predictably become more accurate as more supervision is introduced in the training data.

More importantly, the key distinctive capabilities of our model are demonstrated in \cref{fig:ssl_fami_stra}. As already discussed in \cref{sec:eval_unknown}, the proposed model is capable of detecting complete strangers, and here we see that it correctly predicts that their name is unknown. Furthermore, our model can acknowledge that familiar faces belong to different people, whose names may not be known. Neither of these functionalities is provided by the baselines, as they are limited to the closed-set identification task.



\section{Conclusion}

In this work, we introduced a fully Bayesian treatment of the face identification problem. Each component of our proposed approach was motivated from human intuition about face recognition and tagging in daily social interactions. Our principled identity model can contemplate an unbounded population of identities, accounting for context-specific probabilities of meeting them.

We demonstrated that the proposed identity model can accurately detect when a face is unfamiliar, and is able to incrementally learn to differentiate between new people as they are met in a streaming data scenario. Lastly, we verified that our approach to dealing with sparse name annotations can handle not only acquaintances, whose names are known, but also familiar faces and complete strangers in a unified manner---a functionality unavailable in conventional (semi-) supervised identification methods.

Here we considered a fully supervised context structure. As mentioned in \cref{sec:related_work}, one could imagine an unsupervised approach involving global visual or non-visual signals to drive context inference (\eg global image features, time or GPS coordinates), in addition to extensions to the face model with individual context information (\eg clothing, speech). Yet another interesting research direction is to explicitly consider time dependence, \eg by endowing the sequence of latent contexts with a hidden Markov model-like structure \cite{Torralba2003}.

\subsubsection*{Acknowledgement.}
This work was partly supported by CAPES, Brazil (BEX 1500/2015-05).

\bibliographystyle{splncs04}
\bibliography{references,related}

\appendix
\numberwithin{equation}{section}

\section{Random Measure Interpretation}\label{app:random_measure}

While the exposition in the main text considers the explicit representation of the nonparametric model in terms of weights ($\bpi_0$ and $(\bpi_c)_{c=1}^C$) and atom locations ($(\theta^*_i, y^*_i)_{i=1}^\infty$), here we also provide the interpretation in terms of random measures:
\begin{align}
	\LabelPrior 			&\sim \DP(\lambda, \LabelBase) \\
	G_0 \cond \LabelPrior 	&\sim \DP(\alpha_0, \ObsPrior \otimes \LabelPrior) \\
	G_c \cond G_0 			&\sim \DP(\alpha_c, G_0) \,,	& c &= 1,\dots,C	\label{eq:rm1} \\
	\bomega					&\sim \Dir(\bgamma)									\label{eq:rm2} \\
	c^*_m \cond \bomega		&\sim \Cat(\bomega) \,,			& m &= 1,\dots,M	\label{eq:rm3} \\
	(\theta_n, \tilde{y}_n) \cond f_n = m, \*c^*, (G_c)_c
    						&\sim G_{c^*_m} \,,				& n &= 1,\dots,N	\label{eq:rm4} \\
	\*x_n \cond \theta_n	&\sim \ObsLik(\theta_n) \,,		& n &= 1,\dots,N \\
	y_n \cond \tilde{y}_n, \LabelPrior
    						&\sim \LabelLik(\tilde{y}_n) \,,& n &\in \LabelledIndices \,.
\end{align}
Note that, under this perspective,
\[
	G_0 = \sum_{i=1}^\infty \pi_{0i} \delta_{(\theta^*_i, y^*_i)} \quad \text{and} \quad
    G_c = \sum_{i=1}^\infty \pi_{ci} \delta_{(\theta^*_i, y^*_i)} \,.
\]

Now, if we let $C \to \infty$ as mentioned in footnote \labelcref{foot:unbounded_contexts}, assuming that $\forall c, {\alpha_c = \alpha}$, and $\bgamma = (\tfrac{\gamma_0}{C}, \dots, \tfrac{\gamma_0}{C})$, we obtain the following nested-hierarchical Dirichlet process:
\begin{align}
    Q \cond G_0	&\sim \DP(\gamma_0, \DP(\alpha, G_0)) \\
    G_n \cond Q	&\sim Q \,,		& n &= 1,\dots,N \\
	(\theta_n, \tilde{y}_n) \cond G_n
    			&\sim G_n \,,	& n &= 1,\dots,N \,,
\end{align}
replacing \crefrange{eq:rm1}{eq:rm4}.

\section{Label Likelihood}\label{app:label_lik}

Given the label prior, $\LabelPrior$, we can formulate the following likelihood model:
\begin{equation}\label{eq:label_lik_unobs}
	\LabelLik(\ell \cond y^*_i) = \begin{cases}
		1-\varepsilon \,,	& \ell = y^*_i \\
		\varepsilon \frac{\LabelPrior(\ell)}{1-\LabelPrior(y^*_i)} \,,	& \ell \neq y^*_i
	\end{cases} \,.
\end{equation}

Note that \cref{eq:label_lik_unobs} depends on the unobserved label prior $\LabelPrior$. Fortunately, we are able to marginalise over $\LabelPrior$ to obtain the following convenient result, given $\ell \neq y^*_i$:
\begin{equation}\label{eq:label_expec_prob_wrong}
	\expec*{\frac{\LabelPrior(\ell)}{1-\LabelPrior(y^*_i)}}{\*y^*}
    	= \frac{\LabelPost(\ell \cond \*y^*)}{1-\LabelPost(y^*_i \cond \*y^*)} \,,
\end{equation}
where $\LabelPost$ is defined as in Eq.\ (14) (main paper). This straightforward equivalence arises from the fact that posterior weights in a DP follow a Dirichlet distribution and are therefore neutral: after removing one weight, the proportions between the remaining ones are independent of its value, and they simply follow a Dirichlet distribution with that component discarded.

We can then formulate an alternative likelihood, which depends on $\*y^*$:
\begin{equation}
	\LabelMarg(\ell \cond y^*_i; \*y^*) = \begin{cases}
		1-\varepsilon \,,	& \ell = y^*_i \\
		\varepsilon \frac{\LabelPost(\ell \cond \*y^*)}{1-\LabelPost(y^*_i \cond \*y^*)} \,,	& \ell \neq y^*_i
	\end{cases} \,.
\end{equation}
Although the marginalisation in \cref{eq:label_expec_prob_wrong} breaks the conditional independence of the true component labels, it gives us a simple, tractable form for the likelihoods of observed labels.

The simpler case of uniform label noise, discussed in \cite{Frenay2014}, could not easily be extended to our context with infinite support, as this would result in an improper likelihood $\LabelLik$.

\section{Gibbs Sampler Conditionals}\label{app:gibbs}

Joint posterior:
\[
	\prob{\*z, \*y^*, \btheta^*, \bpi}{\*X, \*y_\LabelledIndices, \*c^*}
\]

\subsection{Global Weights}
As suggested in \cite{Teh2006}, we augment our Markov chain state with the weights of the global DP $G_0$, such that the context DPs $(G_c)_c$ become conditionally independent and can be sampled in parallel:
\begin{equation}
	\bpi_0 = (\pi_{01}, \dots, \pi_{0I}, \pi_0') \cond \*T \sim \Dir(T_{\cdot 1}, \dots, T_{\cdot I}, \alpha_0) \,,
\end{equation}
where $I$ is the current number of distinct identities, $\pi_0'$ is the weight of $G_0$'s base measure ($\pi_0' = \sum_{i=I+1}^\infty \pi_{0i}$) and $T_{\cdot i} = \sum_{c=1}^C T_{ci}$ are auxiliary variables counting the total number of `tables' (context-wise clusters) having `dish' (global cluster) $i$, in the Chinese restaurant analogy \cite{Teh2006}.

Finally, to sample the table counts $\*T$ conditioned on the global weights $\bpi$ and identity and context assignments $\*z$ and $\*c^*$, we use a similar scheme to the one presented in \cite{Dai2015}:
\begin{equation}
	T_{ci} = \sum_{n=1}^{N_{ci}} \indicator*{u_n \leq \frac{\alpha_c \pi_{0i}}{\alpha_c \pi_{0i} + n}} \,,
\end{equation}
where $(u_n)_{n=1}^{N_{ci}}$ are uniformly sampled from $[0,1]$.

\subsection{Identity Assignments}
For the unlabelled instances, we have
\begin{equation}\label{eq:identity_gibbs}
	\prob{z_n}{\*X, \*y_\LabelledIndices, \*z_{-n}, \*c^*, \*y^*, \btheta^*, \bpi_0} \propto \begin{cases}
		\ObsLik(\*x_n \cond \theta^*_i) \, \prob{z_n=i}{\*z_{-n}, \*c^*, \bpi_0} \,, \\
		\ObsPred(\*x_n) \, \prob{z_n \text{ new}}{\*z_{-n}, \*c^*, \bpi_0} \,,
	\end{cases}
\end{equation}
where $\ObsPred(\*x) = \int \ObsLik(\*x \cond \theta) \ObsPrior(\theta) \,\mathrm{d}\theta$, the prior predictive distribution of the observations. The Chinese restaurant franchise conditionals $\prob{z_n}{\*z_{-n}, \*c^*, \bpi_0}$ are given by \cite{Teh2006}
\begin{equation}\label{eq:crf_conditional}
	\prob{z_n = i}{\*z_{-n}, \*c^*, \bpi_0} \propto \begin{cases}
		N_{c_n i}^{-n} + \alpha_{c_n} \pi_{0i} \,,	& N_{c_n i}^{-n} > 0	\\
		\alpha_{c_n} \pi_0' \,,					& i \text{ new}
	\end{cases} \,,
\end{equation}
where $N_{ci} = |\{n : c_n=c \wedge z_n=i\}|$, \ie the number of samples in context $c$ assigned to cluster $i$.

The global weights $\bpi$ are updated whenever an instance gets assigned to a new cluster, by splitting $\pi_0'$ according to the stick-breaking process: sample ${\beta \sim \operatorname{Beta}(1, \alpha_0)}$, then set ${\pi_{0,I+1} \gets \beta \pi_0'}$ and ${\pi_0' \gets (1-\beta) \pi_0'}$ \cite{Teh2006}.

For $n \in \LabelledIndices$, there is an additional term accounting for the likelihood of the observed label:
\begin{multline}
	\prob{z_n}{\*X, \*y_\LabelledIndices, \*z_{-n}, \*c^*, \*y^*, \btheta^*, \bpi_0} \\
		\propto \ObsLik(\*x_n \cond \theta^*_{z_n}) \, \LabelMarg(y_n \cond y^*_{z_n}; \*y^*) \, \prob{z_n}{\*z_{-n}, \*c^*, \bpi_0} \,.
\end{multline}

\subsection{Contexts}
\begin{equation}\label{eq:context_gibbs}
	\prob{c^*_m}{\*X, \*y_\LabelledIndices, \*z, \*c^*_{-m}, \*y^*, \btheta^*, \bpi_0}
		\propto \prob{\*z_{\FrameIndices_m}}{\*z_{-\FrameIndices_m}, \*c^*, \bpi_0} \, \prob{c^*_m}{\*c^*_{-m}}
\end{equation}

The context posterior predictive distribution is
\begin{equation}
	\prob{c^*_m = c}{\*c^*_{-m}} \propto \frac{\gamma_0}{C} + M_c^{-m} \,,
\end{equation}
where $M_c^{-m}$ is the number of frames assigned to context $c$, excluding frame $m$.

The conditional distribution for the identities in frame $m$ can be computed via sequential application of \cref{eq:crf_conditional}:
\begin{equation}
	\prob{\*z_{\FrameIndices_m}}{\*z_{-\FrameIndices_m}, \*c^*, \bpi_0} = \prod_{r=1}^{|\FrameIndices_m|} \prob{z_{\FrameIndices_m}^{(r)}}{\*z_{\FrameIndices_m}^{(<r)}, \*z_{-\FrameIndices_m}, \*c^*, \bpi_0} \,,
\end{equation}
where $r$ indexes observations within each single frame. Note that, due to exchangeability of the HDP, the order of iteration of $r$ is inconsequential.

\subsection{Labels}

Let $J_\ell^{-i} = |\{j : y^*_j = \ell \wedge j \neq i\}|$, the number of identities with label $\ell$ excluding identity $i$, and $\LabelledIndices^{(i)} = \{n \in \LabelledIndices : z_n = i\}$, the indices of labelled observations assigned to identity $i$. We can then write the Gibbs identity label predictive as
\begin{equation}
	\LabelPost(y^*_i \cond \*y^*_{-i}) = \frac{1}{\lambda + I - 1} \begin{cases}
		\lambda \LabelBase(\ell) + J_\ell^{-i} \,,		& y^*_i = \ell \in \LabelSetKnown \\
        \lambda (1 - \LabelBase(\LabelSetKnown)) \,,	& y^*_i \notin \LabelSetKnown
	\end{cases} \,,
\end{equation}
where $\LabelSetKnown$ is the set of all known labels, whether allocated to components or not. Additionally, recall that the label likelihood is
\begin{equation}
	\LabelMarg(y \cond \ell; \*y^*) = (1 - \varepsilon)^{\indicator{y = \ell}} \left[ \varepsilon \frac{\LabelPost(y \cond \*y^*)}{1 - \LabelPost(\ell \cond \*y^*)} \right]^{\indicator{y \neq \ell}} \,.
\end{equation}

The probability of assigning a label $\ell$ to identity $i$, given the remaining identity labels, can be computed as
\begin{equation}
\begin{split}
	\LHS \prob{y^*_i = \ell}{\*y_\LabelledIndices, \*z_\LabelledIndices, \*y^*_{-i}}
    	\propto \LabelPost(\ell \cond \*y^*_{-i}) \prod_{n \in \LabelledIndices^{(i)}} \LabelMarg(y \cond \ell; \*y^*) \\
    	&\propto \frac{\lambda \LabelBase(\ell) + J_\ell^{-i}}{\lambda + I - 1}
    		(1 - \varepsilon)^{|\LabelledIndices_\ell^{(i)}|}
        	\prod_{k \in \LabelSetKnown \setminus \{\ell\}} \left[ \frac{\varepsilon (\lambda \LabelBase(k) + J_k)}{\lambda + I - (\lambda \LabelBase(\ell) + J_\ell)} \right]^{|\LabelledIndices_k^{(i)}|} \,,
\end{split}
\end{equation}
where $\LabelledIndices_\ell^{(i)} = \{n \in \LabelledIndices^{(i)} : y_n = \ell\}$.

First, let us consider the probability of assigning a \emph{known} label to identity $i$:
\begin{align*}
	\LHS \prob{y^*_i = \ell \in \LabelSetKnown}{\*y_\LabelledIndices, \*z_\LabelledIndices, \*y^*_{-i}} \\
    	&\propto \frac{\lambda \LabelBase(\ell) + J_\ell^{-i}}{\lambda + I - 1}
	    	(1 - \varepsilon)^{|\LabelledIndices_\ell^{(i)}|}
	        \left[ \frac{\varepsilon (\lambda \LabelBase(\ell) + J_\ell)}{\lambda + I - (\lambda \LabelBase(\ell) + J_\ell)} \right]^{-|\LabelledIndices_\ell^{(i)}|} \\
            & \qquad \times \prod_{k \in \LabelSetKnown} \left[ \frac{\varepsilon (\lambda \LabelBase(k) + J_k)}{\lambda + I - (\lambda \LabelBase(\ell) + J_\ell)} \right]^{|\LabelledIndices_k^{(i)}|} \\
		&\approx \frac{\lambda \LabelBase(\ell) + J_\ell^{-i}}{\lambda + I - 1}
	    	(1 - \varepsilon)^{|\LabelledIndices_\ell^{(i)}|}
	        \left[ \frac{\varepsilon (\lambda \LabelBase(\ell) + J_\ell)}{\lambda + I - J_\ell} \right]^{-|\LabelledIndices_\ell^{(i)}|}
            \prod_{k \in \LabelSetKnown} \left[ \frac{\varepsilon (\lambda \LabelBase(k) + J_k)}{\lambda + I - J_\ell} \right]^{|\LabelledIndices_k^{(i)}|} \\
		&= \frac{\lambda \LabelBase(\ell) + J_\ell^{-i}}{\lambda + I - 1}
	    	\left[ \frac{(1 - \varepsilon)(\lambda + I - J_\ell)}{\varepsilon (\lambda \LabelBase(\ell) + J_\ell)} \right]^{|\LabelledIndices_\ell^{(i)}|}
	        \frac{\prod_{k \in \LabelSetKnown} [\varepsilon (\lambda \LabelBase(k) + J_k)]^{|\LabelledIndices_k^{(i)}|}}{(\lambda + I - J_\ell)^{|\LabelledIndices^{(i)}|}} \\
		&\propto \frac{\lambda \LabelBase(\ell) + J_\ell^{-i}}{(\lambda + I - J_\ell)^{|\LabelledIndices^{(i)}|}}
	    	\left[ \frac{(1 - \varepsilon) (\lambda + I - J_\ell)}{\varepsilon (\lambda \LabelBase(\ell) + J_\ell)} \right]^{|\LabelledIndices_\ell^{(i)}|} \,.
        \numberthis\label{eq:prob_label_known}
\end{align*}
where the approximation assumes that $\lambda \LabelBase(\ell) \ll \lambda + I - J_\ell$, $\forall \ell$, which is generally the case for sensible choices of $\lambda$ and $\LabelBase$.

We can analogously estimate the probability of assigning an \emph{unknown} label to an identity as follows:
\begin{align*}
	\prob{y^*_i \notin \LabelSetKnown}{\*y_\LabelledIndices, \*z_\LabelledIndices, \*y^*_{-i}} &= \sum_{\ell \notin \LabelSetKnown} \prob{y^*_i = \ell}{\*y_\LabelledIndices, \*z_\LabelledIndices, \*y^*_{-i}} \\
    	&\propto \sum_{\ell \notin \LabelSetKnown} \frac{\lambda \LabelBase(\ell)}{\lambda + I - 1}
        	\prod_{k \in \LabelSetKnown} \left[ \frac{\varepsilon (\lambda \LabelBase(k) + J_k)}{\lambda + I - \lambda \LabelBase(\ell)} \right]^{|\LabelledIndices_k^{(i)}|} \\
		&\approx \sum_{\ell \notin \LabelSetKnown} \frac{\lambda \LabelBase(\ell)}{\lambda + I - 1}
        	\prod_{k \in \LabelSetKnown} \left[ \frac{\varepsilon (\lambda \LabelBase(k) + J_k)}{\lambda + I} \right]^{|\LabelledIndices_k^{(i)}|} \\
		&= \frac{\lambda (1 - \LabelBase(\LabelSetKnown))}{\lambda + I - 1}
        	\prod_{k \in \LabelSetKnown} \left[ \frac{\varepsilon (\lambda \LabelBase(k) + J_k)}{\lambda + I} \right]^{|\LabelledIndices_k^{(i)}|} \\
		&\propto \frac{\lambda (1 - \LabelBase(\LabelSetKnown))}{(\lambda + I)^{|\LabelledIndices^{(i)}|}} \,,
        \numberthis\label{eq:prob_label_unknown}
\end{align*}
noting that $J_\ell = J_\ell^{-i} = |\LabelledIndices_\ell^{(i)}| = 0$ for $\ell \notin \LabelSetKnown$ and using a similar approximation as in \cref{eq:prob_label_known}.

Finally, combining \cref{eq:prob_label_known,eq:prob_label_unknown}, we can summarise
\newcommand{\approxpropto}{\overset{\sim}{\propto}}
\begin{multline}
	\prob{y^*_i}{\*X, \*y_\LabelledIndices, \*z, \*c^*, \*y^*_{-i}, \btheta^*, \bpi_0} \\
		\approxpropto \begin{cases}
			\dfrac{\lambda \LabelBase(\ell) + J_\ell^{-i}}{(\lambda + I - J_\ell)^{|\LabelledIndices^{(i)}|}}
	    	\left[ \dfrac{(1 - \varepsilon) (\lambda + I - J_\ell)}{\varepsilon (\lambda \LabelBase(\ell) + J_\ell)} \right]^{|\LabelledIndices_\ell^{(i)}|},
				& y^*_i = \ell \in \LabelSetKnown \\[2.5ex]
			\dfrac{\lambda (1 - \LabelBase(\LabelSetKnown))}{(\lambda + I)^{|\LabelledIndices^{(i)}|}},
				& y^*_i \notin \LabelSetKnown
		\end{cases} \,,
\end{multline}
where $\approxpropto$ means \emph{approximately proportional to}.

\subsection{Face Feature Parameters}
\begin{equation}
	\prob{\theta^*_i}{\*X, \*y_\LabelledIndices, \*z, \*c^*, \*y^*, \btheta^*_{-i}, \bpi_0}
		\propto \ObsPrior(\theta^*_i) \prod_{\mathclap{n:z_n=i}} \ObsLik(\*x_n \cond \theta^*_i ) \,,
\end{equation}
which will be analytically tractable if $\ObsLik$ and $\ObsPrior$ are a conjugate pair.

\end{document}